\documentclass[10pt,twocolumn,letterpaper]{article}

\usepackage{cvpr}
\usepackage{times}
\usepackage{epsfig}
\usepackage{graphicx}
\usepackage{amsmath}
\usepackage{amssymb}

\usepackage{algorithmic}
\usepackage{algorithm} 
\usepackage{enumitem}
\usepackage{caption}
\usepackage{subcaption}
\usepackage[breaklinks=true,bookmarks=false]{hyperref}

\cvprfinalcopy 

\def\cvprPaperID{0008} 

\ifcvprfinal\fi

\DeclareMathOperator*{\argmax}{arg\,max}
\begin{document}

\title{Viewpoint Optimization for Autonomous Strawberry Harvesting \\ with Deep Reinforcement learning}

\author{Jonathon Sather and Xiaozheng Jane Zhang\\
California Polytechnic State University\\
San Luis Obispo, CA\\
{\tt\small \{jsather, jzhang\}@calpoly.edu}
}

\maketitle

\begin{abstract}
Autonomous harvesting may provide a viable solution to mounting labor pressures in the United States's strawberry industry. However, due to bottlenecks in machine perception and economic viability, a profitable and commercially adopted strawberry harvesting system remains elusive. In this research, we explore the feasibility of using deep reinforcement learning to overcome these bottlenecks and develop a practical algorithm to address the sub-objective of viewpoint optimization, or the development of a control policy to direct a camera to favorable vantage points for autonomous harvesting. We evaluate the algorithm's performance in a custom, open-source simulated environment and observe encouraging results. Our trained agent yields 8.7 times higher returns than random actions and 8.8 percent faster exploration than our best baseline policy, which uses visual servoing. Visual investigation shows the agent is able to fixate on favorable viewpoints, despite having no explicit means to propagate information through time. Overall, we conclude that deep reinforcement learning is a promising area of research to advance the state of the art in autonomous strawberry harvesting.
\end{abstract}

\section{Introduction}\label{sec:intro}

Driving down Highway 101 through Santa Maria, California in the springtime, it is hard to ignore the abundance of strawberry fields on both sides of the highway. Each field is populated by dozens of seasonal laborers inching their way down the rows and stooping to manually collect ripe berries -- a practice that has been largely unchanged for the past 700 years~\cite{grubinger_2012}. Recently, the heavy reliance on human labor has become problematic for the California strawberry industry, as an aging and ever-shrinking workforce continues to drive up costs~\cite{mohan_2017}.

A promising idea to address these mounting labor pressures is to use autonomous harvesting to fill the roles of human pickers. In such a system, robots navigate the strawberry field and remove ripe strawberries, eliminating the need for manual labor. Although many researchers are actively working to improve autonomous harvesting technology, a profitable and commercially adopted strawberry harvesting system remains elusive~\cite{charles_2018}.  
Strawberry plants present several difficulties for autonomous harvesting systems, including high levels of occlusion, lighting variation, and easily-bruised fruit. Deep reinforcement learning may provide the machinery for a harvesting agent to develop advanced control policies that capture the complex relationships required to reason about the unstructured harvesting environment. However, deep reinforcement learning systems are historically fragile and have seen limited success in real-world settings~\cite{deep_17}. Therefore, it is not obvious whether such an approach can be successfully adapted to the autonomous harvesting domain. 

\begin{figure}[!t]
\centering
\includegraphics[width=3.25in]{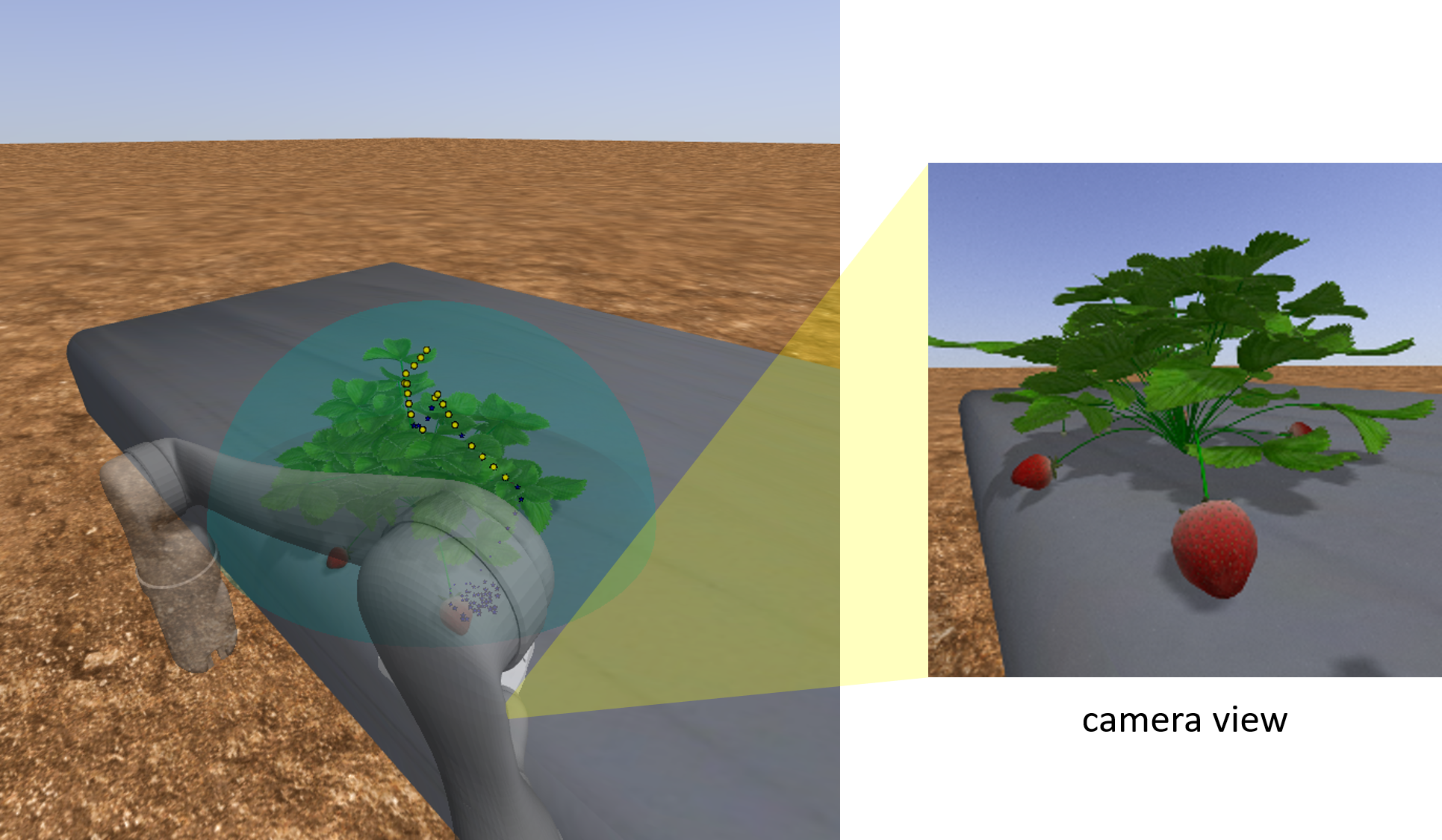}
\caption{Trajectory visualization in the simulated environment. Reinforcement learning is used to train a high degree-of-freedom robot to position its camera towards favorable vantage points for autonomous harvesting.}
\label{fig:viewpoint_opt}
\end{figure}

We narrow the scope from the full harvesting process to the task of viewpoint optimization, or the development of a control policy to direct a camera to favorable vantage points for autonomous harvesting (Figure~\ref{fig:viewpoint_opt}). Viewpoint optimization is a seldom-addressed paradigm that could be prepended to an autonomous harvesting pipeline to improve speed and accuracy on the remaining harvesting steps, \eg object detection, pose determination, path planning, and berry removal~\cite{DefterliReview}. More importantly, viewpoint optimization encapsulates many of the perception bottlenecks present in the full harvesting task, while significantly reducing the overall task complexity. For these reasons, we believe that a successful application of deep reinforcement learning to the viewpoint optimization objective will provide considerable insight on the framework's potential utility in the full harvesting process. 

\subsection{General Approach}
We develop a novel application of Deep Deterministic Policy Gradients (DDPG)~\cite{lillicrap2015continuous} that leverages a pretrained object detector to provide reward feedback and facilitate autonomous training. To collect training images for the object detector, we propose a labor efficient data collection procedure, which makes the entire process require minimal human interaction in a real-world setting. We then create a multi-purpose simulated harvesting environment\footnote{\url{https://github.com/jsather/harvester-sim}} using ROS~\cite{ros} and Gazebo~\cite{gazebo} which we use to train and test all components of our viewpoint optimization algorithm. Our results show that deep reinforcement learning is a promising area of research to advance the state of the art in autonomous strawberry harvesting. 

\section{Related Work}

To the best of our knowledge, this work is the first research exploring viewpoint optimization via reinforcement learning for autonomous harvesting applications (any crop). A related idea is visual servoing~\cite{chaumette2006visual}, which is a class of control algorithms acting on image features that has been applied in the autonomous harvesting domain. Unlike our method, visual servoing typically involves hand-specifying features, whereas our algorithm learns a control policy with a data-driven approach. In~\cite{MEHTA2014146}, Mehta and Burks implement visual servoing by means of two cameras: one in the hand of a citrus harvesting robot and one stationary camera with a wide field of view. The feedback from the cameras is then used to create a perspective image and guide the robotic manipulator towards an artificial citrus fruit. One of the main limitations of this approach is that it requires the target fruit to be visible by the fixed camera, which cannot be guaranteed in unstructured environments.

The main algorithms used in this research are Deep Deterministic Policy Gradients (DDPG)~\cite{lillicrap2015continuous} and You Only Look Once, Version 2 (YOLOv2)~\cite{redmon2016yolo9000}. DDPG is an off-policy, actor-critic deep reinforcement learning algorithm that is used as the underlying machinery for the viewpoint optimization problem. YOLOv2 is a single-shot object detection algorithm using convolutional neural networks. It is capable of outputting labeled bounding boxes at above real-time speeds using consumer-level hardware.  In this research, we train YOLOv2 to detect strawberries and use its output as a feedback mechanism during the reinforcement learning process. 

\section{Preliminaries}

Reinforcement learning problems are often modeled as a Markov Decision Process (MDP), which describes a discrete-time, stochastic environment with a decision-making agent~\cite{Bel}. MDPs are defined by the 5-tuple $\{S, A, P, R, \gamma\}$, where:
\begin{enumerate}[nolistsep] 
    \item $s_t \in S$ is the set of all states, each $s_t$ containing all relevant information about the environment at time $t$.
    \item $a_t \in A$ is the set of all possible actions in the environment at time $t$.
    \item $P(s_{t+1}|s_t,a_t)$ is the state-transition probability for state $s_{t+1}$ given state $s_t$ and action $a_t$.
    \item $R(r_t | s_t, a_t)$ is the reward probability for $r_t$ given state $s_t$ and action $a_t$.
    \item $\gamma \in [0,1)$ is the discount factor, which is used to geometrically decay the value of future rewards and often aids in algorithm convergence~\cite{DBLP:books/lib/SuttonB98}.
\end{enumerate}

The resulting process is characterized by a cyclic interplay between an agent and its environment. At timestep $t$, an agent observes state $s_t$ and subsequently takes action $a_t$ according to its policy $\pi(a_t|s_t)$. The agent then arrives at state $s_{t+1}$ through the environment's dynamics $P(s_{t+1}|s_t,a_t)$ and receives scalar reward $r_t$ according to $R(r_t | s_t, a_t)$. This process repeats, now from $s_{t+1}$, until a termination criterion is reached. 

A roll-out of states and actions from initial state $s_0$ until termination is called a trajectory, denoted $\tau = (s_0, a_0, s_1, a_1, \ldots{},s_{T-1}, a_{T-1})$.  We denote the discounted return for a given trajectory as $\mathcal{R} = \sum^{T-1}_{t=0}\gamma^t r_{t}$. For our problem, the goal of reinforcement learning is discover an optimal policy $\pi^*$ that maximizes expected return under trajectory distribution $p_{\pi}(\tau)$. Formally: 
\begin{equation} 
    \pi^* = \argmax_{\pi}\mathbb{E}_{\tau \sim p_{\pi}(\tau)}[\mathcal{R}]\label{eq:rl_objective}
\end{equation}

In most practical settings, it is infeasible to explicitly represent the policy or expected return at each $s\in S$ and $a \in A$, so it is common to use function approximators to characterize these quantities at regions of interest. DDPG uses neural networks to approximate policy $\pi_{\phi}$ and state-action value function $Q_{\theta}(s_t, a_t) \approx \mathbb{E}_{\tau \sim p_{\pi}(\tau)}[R|s_t, a_t]$. During each training iteration, network weights are jointly updated to maximize the reinforcement learning objective using DDPG, an implementation of Deterministic Policy Gradients (DPG)~\cite{Silver2014DeterministicPG} adapted to work deep neural networks. 

Specifically, $Q_{\theta}$ is updated using a variant of fitted Q-iteration with deterministic policy $\pi_{\phi}$ used to approximate $\argmax_{a}Q_{\theta}(s_t, a)$. $\pi_{\phi}$ is updated using the deterministic policy gradient to maximize $Q_{\theta}(s_t, \pi_{\phi}(a_t|s_t))$. Lillicrap \textit{et al.}\ leverage techniques inspired by Deep Q Networks (DQN)~\cite{Mnih2013PlayingAW}, such as experience replay and target networks $Q_{\theta'}, \pi_{\phi'}$, to stabilize training. This gives way to the following off-policy updates: 
\begin{align}
    y_i &= r_i + \gamma Q_{\theta'}(s_{i+1}, \pi_{\phi'}(s_i)) \\
    \theta &\gets \theta - \epsilon_{\theta} \sum_i \nabla_{\theta} || y_i - Q_{\theta}(s_i, a_i) ||^2 \\
    \phi &\gets \phi + \epsilon_{\phi} \sum_i \nabla_{a}Q_{\theta}(s_i, \pi(s_i)) \nabla_{\phi}\pi(s_i)
\end{align}
where subscript $i$ is used to reference elements of the current training batch. Further details can be found in~\cite{lillicrap2015continuous}. 

\subsection{A Note on Partial Observability}
It is worth noting that while the the forgoing discussion assumes the reinforcement learning problem could be framed as a Markov Decision Process, this is not always the case. In many real-world problems, such as our formulation of viewpoint optimization, each observation taken in the environment may not contain all relevant information to maximize expected return. In this case, the environment is said to be partially observed, which often warrants a generalization of the MDP framework~\cite{KAELBLING199899}. Despite this, we use algorithms designed for fully observed MDPs in a partially observed setting. Although we lose some theoretical support, this simplifies our implementation while still facilitating meaningful insights for reinforcement learning's promise in autonomous harvesting.

\section{Method}\label{sec:method}

\subsection{System Setup}
\begin{figure}[!t]
\centering
\includegraphics[width=0.8\linewidth]{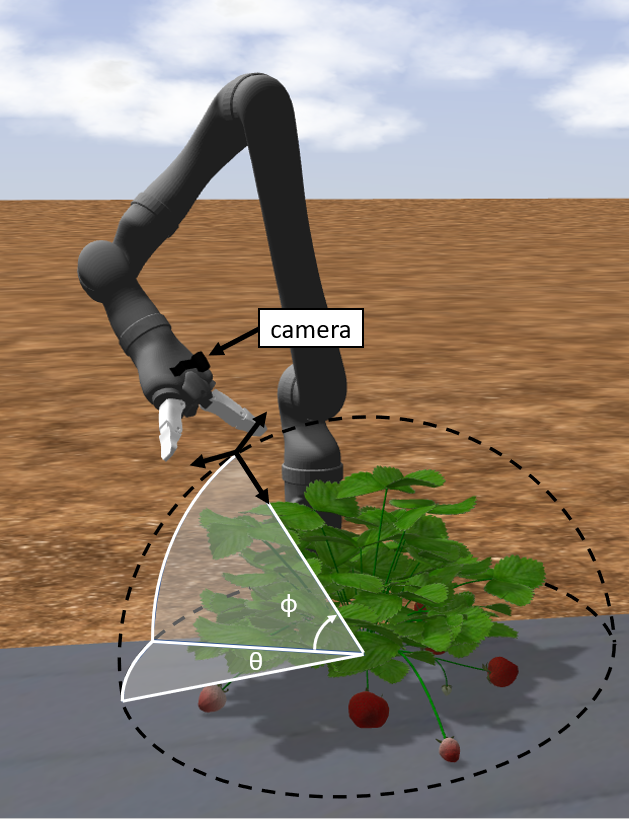}
\caption{Visualization of the state and action space in the simulated harvesting environment.}
\label{fig:mdp_visual}
\end{figure}

We define our environment to consist of a high degree-of-freedom articulated robot with an RGB camera mounted to its end effector, positioned over an outdoor strawberry plant. To avoid plant collisions, we constrain the end-effector to move on a hemisphere of a fixed radius above the strawberry plant as shown in in Figure~\ref{fig:mdp_visual}. This allows us to define its position by two rotation angles $j = (\theta, \phi)$. The set of all reachable positions combined with all possible camera images defines our state space, $S=\{\mathcal{J}, \mathcal{O}\}$. We define our action space as incremental positions on this hemisphere $a = (\Delta\theta, \Delta\phi)$ and use off-the-shelf trajectory planning software and proportional-integral-derivative (PID) joint controllers to move between states.

At each state, we use confidence values from a pretrained strawberry detector and return positive reward if ripe strawberry confidence is above a given threshold. We also penalize actions that lead to unreachable positions and impose a light ``existence penalty" to encourage exploration when no ripe strawberries are visible. Intuitively, this reward scheme encourages the agent to move to viewpoints that yield high probability of ripe strawberry detection, which we assume correspond to favorable viewpoints for the harvesting process. Note that even if this assumption is not fully met, we can still gain insight from the agent's performance, as the task nevertheless requires the agent to overcome common perception challenges in a stochastic harvesting environment. Formally, the reward scheme is specified as follows:
\begin{equation}\label{eq:reward_scheme}
    R(s_t) = \left\{
    \begin{array}{ll}
          R_{invalid} & (\theta_t, \phi_t) \notin \mathcal{J} \\
          R_{detect} & P_{max,t} \geq P_{thresh} \\
          R_{exist} & otherwise
    \end{array} 
    \right.        
\end{equation}
where $P_{max,t}$ and $P_{thresh}$ correspond to the maximum ripe strawberry confidence output by the detector at timestep $t$ and the minimum confidence threshold, respectively. In our experiments, we use $R_{detect} = 1.0$, $R_{invalid} = -1.0$, $R_{exist} = -0.1$, and $P_{thresh} = 0.6$.

One of the primary benefits of our reward specification is that it does not require a human-in-the-loop to provide external feedback. This makes it convenient to train the harvester for extended intervals in environments where policy execution is expensive, \eg the real world. A practical limitation of this approach is that we are relying on a strawberry detector to produce ``correct" annotations without reference to ground-truth values, which may allow the agent to exploit false positive rewards. To mitigate this limitation, we ensure the strawberry detector predicts robust and high-precision bounding boxes before integrating it into the reinforcement learning algorithm.

\subsection{Detector}

While our reward scheme eliminates the need for human feedback during reinforcement learning, it introduces the new burden of collecting and annotating images for pretraining the strawberry detector. To combat this burden, we develop a labor-efficient data collection procedure that leverages the fact that multiple training images can be annotated per plant given knowledge of its berries' locations in 3D space. This procure is outlined in Algorithm~\ref{alg:get_detector_data}. Note that we collect data and train our detector for both ripe and unripe strawberries, despite only using ripe detections in our reward scheme. This is done to increase the detector's ability to discriminate between the two classes and reduce the frequency of false positive detections.  

In a real-world harvesting context, the ground truth 3D strawberry labels in Algorithm~\ref{alg:get_detector_data} can be obtained by kinesthetically guiding the robot's end effector around each strawberry to log it's location and approximate diameter. In the simulated environment, we replace this step by directly using the ground truth poses and sizes of each strawberry. To simplify the mapping from 3D pose to bounding box, we approximate each strawberry as a perfect sphere. 

During the annotation step, we account for occlusions by rejecting bounding boxes with insufficient pixels within an empirically-determined hue range. We find this occlusion heuristic results in satisfactory annotations on our simulated data without requiring manual labelling. We note that a more sophisticated occlusion removal strategy may be needed in a real-world setting.  

We use Algorithm~\ref{alg:get_detector_data} to collect over 7000 images in the simulated environment, designating 80\% of dataset for training/validation and the remaining 20\% for testing. We then train YOLOv2 for 50,000 epochs using the Darknet framework~\cite{darknet13} with default hyperparameters as specified in~\cite{redmon2016yolo9000}. The trained detector is able to perform real-time detections ($> 30$ frames per second) using a Tesla K80 GPU.

\subsection{Reinforcement Learning}

We frame the viewpoint optimization problem as an episodic MDP, where each episode the agent is spawned at fixed location $j_0 = (\theta_0, \phi_0)$ relative to an arbitrary strawberry plant. In a real-world harvesting environment, this could be implemented by mounting the manipulator to an unmanned ground vehicle and navigating from plant to plant between episodes. The entire process can the be executed without human intervention, provided the robot is equipped with appropriate vision software for autonomous navigation and safe operation. In the simulated environment, this process is mimicked using a generative strawberry plant model to initialize a new plant configuration at the beginning of each episode. 

\begin{algorithm}[!t]
\caption{Detector Data Collection}\label{alg:get_detector_data}
\begin{algorithmic}[]
    \STATE initialize dataset $\mathcal{D}$
    \FOR{i = 0, 1, 2, \ldots{} until $num\_plants$}
        \STATE move to next plant
        \STATE record location, size, and label of each strawberry 
        \FOR{k = 0, 1, 2, \ldots{} until $num\_views$}
            \STATE uniformly sample position $j_k$ in workspace $\mathcal{J}$
            \STATE actuate camera to $j_k$ and capture image $o_k$
            \STATE initialize set of valid bounding boxes $\mathcal{B}_k$
            \FOR{each strawberry}
                \STATE project 3D pose to bounding box $b$ 
                \IF{$b$ not occluded}
                    \STATE $\mathcal{B}_k \gets \mathcal{B}_k \cup \{b\}$
                \ENDIF
            \ENDFOR
            \STATE $\mathcal{D} \gets \mathcal{D} \cup \{\mathcal{B}_k, o_k\}$
        \ENDFOR
    \ENDFOR 
    \RETURN $\mathcal{D}$
\end{algorithmic}
\end{algorithm}
We use DDPG to train the harvesting agent for viewpoint optimization, as its off-policy nature allows for sample efficient training updates. Additionally, DDPG has seen success using convolutional neural networks to process raw pixel inputs~\cite{lillicrap2015continuous}. Our implementation is very similar to~\cite{lillicrap2015continuous}, except we perform policy execution and training updates in parallel to maximize sample efficiency (inspired by~\cite{DBLP:journals/corr/VecerikHSWPPHRL17}).

For the both the actor $\pi_{\phi}$ and critic $Q_{\theta}$, we use five $3 \times 3 \times 32$ convolutional layers with stride 2 to process the $800 \times 800 \times 3$ raw pixel input. In the actor network, this is followed by two 200-neuron fully connected layers with tanh activations. The critic network has a similar structure, except the first fully connected layer is concatenated with the action input, and the final (scalar) output is a linear activation. The networks use the same uniform initialization scheme and weight regularization as in~\cite{lillicrap2015continuous}. Training is performed in batch sizes of 16 and updates are performed with Adam optimizer~\cite{adam_opt} using learning rates of $1 \times 10^{-4}$ and $1 \times 10^{-3}$ for the actor and critic, respectively, and default values $\beta_1 = 0.9$, $\beta_2 = 0.999$, $\hat{\epsilon} = 1 \times 10^{-8}$ from the paper. The target networks are updated using Polyak Averaging~\cite{Polyak:1992:ASA:131092.131098} with mixing parameter $\tau=1 \times 10^{-3}$. We implement the network architectures using TensorFlow~\cite{tensorflow2015-whitepaper} and train for 10,000 episodes in the simulated environment. On a physical system, this equates to approximately one week of training.

\subsection{Simulated Environment}

We create a simulated environment using Robot Operating System (ROS)~\cite{ros} and Gazebo~\cite{gazebo} with the goal of being sufficiently realistic so that results in the simulated environment are indicative of performance in real-world settings. The virtual world mimics an open strawberry field and consists of a section of a bed with dirt surroundings and a randomly-generated strawberry model. The strawberry model incorporates real-world perception challenges such as heavy occlusion, complex textures, shadows, and stochasticity. Within the world, we place a ``floating arm" harvester centered about the strawberry plant. The arm is modelled after the JACO by Kinova Robotics~\cite{kinova_robotics}. 

Plant models are generated using using Gazebo's native SDF file format with embedded Ruby~\cite{erb}. The algorithm creates plants based on a simple structural model in which various parameters, such as berry pose, mesh, and ripeness, are sampled to create unique plant configurations on the fly. All of the meshes are custom made, with the exception of the strawberry flesh, which uses down-sampled meshes from the UC Davis Strawberry Database~\cite{ucdavis_database}.

\begin{figure}[!t]
\centering
\hspace*{-5mm}
\includegraphics[width=0.95\linewidth]{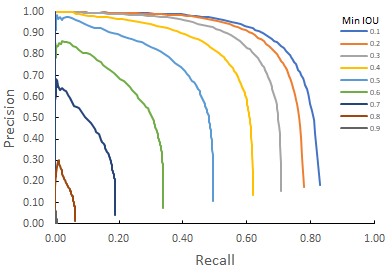}
\caption{Precision-recall curve on the test set of 2000 held-out images. Each color represents a different minimum IOU for positive detection.}
\label{fig:PR_curve}
\end{figure}

\section{Results}\label{sec:results}

\subsection{Detector Evaluation}

We evaluate the learned strawberry detector on 2000 held-out images to generate the precision-recall plot in Figure~\ref{fig:PR_curve}. The plot shows the relationship between precision and recall as the YOLOv2 minimum confidence is varied from 0.01 to 1.0 for 9 intersection-over-union (IOU) thresholds. Looking at the ``traditional" IOU metric of 0.5, we see that there is a gradual linear decrease in precision with a large increase in recall as the threshold is lowered from 0.75 to 0.5. Lowering the threshold further results in a sharp decrease in precision and asymptotic recall to 0.5.

In reinforcement learning, it is important to have a strong reward signal so that the agent can understand the consequences of its actions. As such, we err on the side of high precision/low recall and select $R_{thresh}=0.6$ as our reward threshold, corresponding to a precision of 0.9 and a recall of less than 0.2. While such a low recall may initially raise some red flags, it is important to note that its implications largely depend on the nature of the strawberries that are ignored. For this, we turn to a visual analysis. 

\begin{figure}[!b]
\centering
\includegraphics[width=0.8\linewidth]{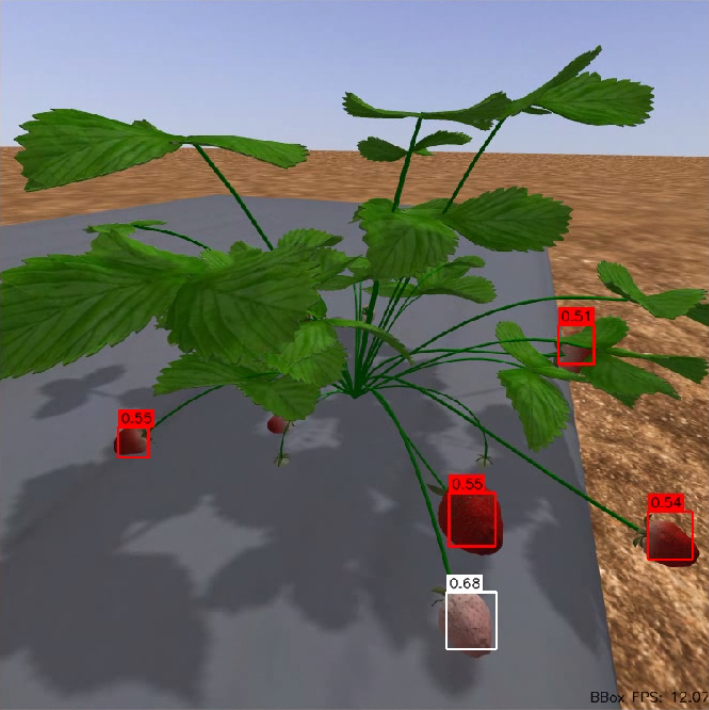}
\caption{Screenshot of the pretrained object detector annotating the simulated camera feed. The detector appears to prefer strawberries that are larger and unoccluded.}
\label{fig:detector_example}
\end{figure}

Looking at the behavior of the detector frame-by-frame, we note its performance is relatively consistent across the camera images. It appears to prefer strawberries that are closer to the camera and larger, with more red flesh of the strawberry showing corresponding to higher confidence values. The strawberries that are missed by the detector are typically smaller or heavily occluded. Therefore, as a feedback mechanism for viewpoint optimization, the biases exhibited by the detector are likely beneficial. An example annotation is shown in Figure~\ref{fig:detector_example}. 

\subsection{Policy Performance}

To assess the performance of the learned policy, we compare several performance metrics versus five baseline policies and a ``hybrid" policy. These policies are listed below in order of increasing complexity. 

\begin{enumerate}[nolistsep]
    \item Random policy, $\pi_1$: At each timestep, the agent takes a uniformly sampled action in $\Delta \theta$ and $\Delta \phi$. 
    
    \item Random policy with boundary awareness, $\pi_2$: At each timestep, the agent takes a uniformly sampled action in $\Delta \theta$ and $\Delta \phi$, taking opposite actions to stay in bounds as needed. 
    
    \item Downward heuristic with boundary awareness, $\pi_3$: Given threshold $\phi^* \in (0, \frac{\pi}{2})$. At each timestep, the policy selects a downward action along the hemisphere until it reaches threshold $\phi^*$. Once it crosses the threshold, the policy takes random actions in accordance with $\pi_2$. 

    \item Frozen detector with downward heuristic and boundary awareness, $\pi_4$: At each timestep, the policy first runs the pretrained strawberry detector on observation $o_t$, and obtains coordinates $(x_t, y_t)$ of the most confident ripe detection in the image frame. If a ripe strawberry is detected, the policy outputs zero action. Otherwise, the policy moves in accordance with $\pi_3$. 
    
    \item Proportional detector with downward heuristic and boundary awareness, $\pi_5$: At each timestep, the policy first runs the pretrained strawberry detector on observation $o_t$, and obtains coordinates $(x_t, y_t)$ of the most confident ripe detection in the image frame. If a ripe strawberry is detected, the policy outputs action proportional to the direction of its bounding box in the image plane. Otherwise, the policy moves in accordance with $\pi_3$. 

    \item Hybrid policy: Given threshold $\phi^* \in (0, \frac{\pi}{2})$. At each timestep, the policy selects a downward action along the hemisphere until it reaches threshold $\phi^*$. Once it crosses the threshold, the agent follows the learned policy (DDPG), taking opposite actions to stay in-bounds as needed.
\end{enumerate}

\begin{figure}[!t]
\centering
\includegraphics[width=3.25in]{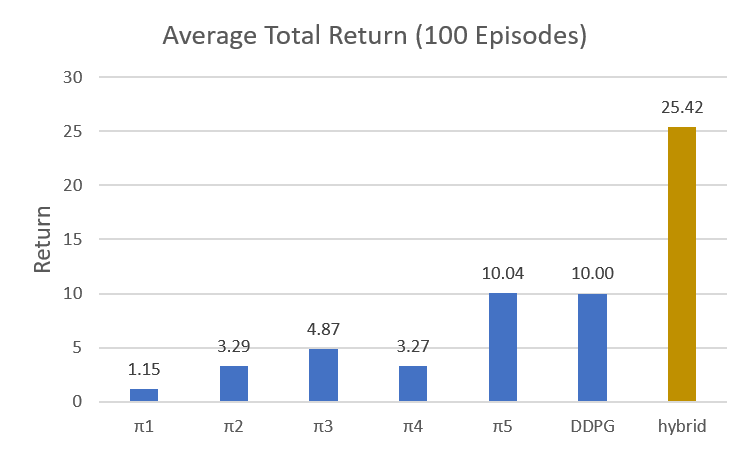} 
\caption{\label{fig:avg_return} Mean return over 100 episodes for each policy. The hybrid policy, which combines the learned policy with hard-coded rules, outperforms all by a large margin.} 
\end{figure} 

Note that for the detector-based policies we use a bounding box threshold of 0.5 to represent a positive detection, instead of 0.6 used to derive the reward values. This prevents the baseline detectors from having ``insider knowledge" of the reward scheme and makes the resulting comparison less biased. 

\begin{figure}[!t] 
\centering
\includegraphics[width=3.25in]{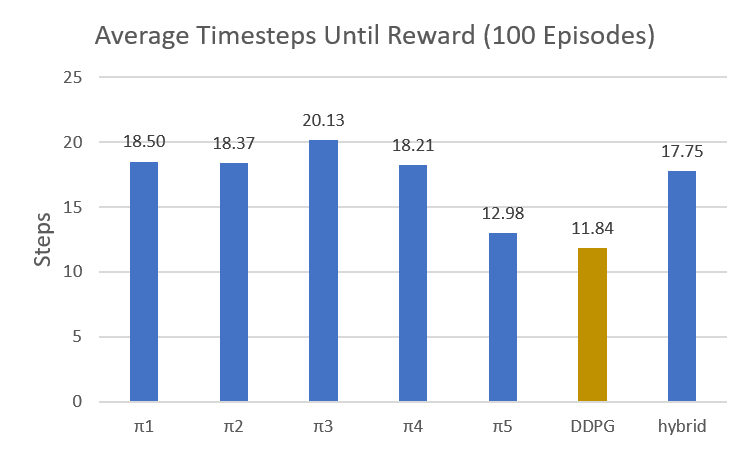}
\caption{\label{fig:timesteps_reward} Mean timesteps until first reward over 100 episodes, excluding episodes without a reward. Under this metric, the hybrid policy causes a decrease in performance.}
\end{figure}

We run each of the policies for 100 episodes on previously unseen strawberry plants, recording the reward at each timestep and number of steps per episode. Using this information, we calculate the mean return for each episode and the number of timesteps until first reward, excluding trials without a reward. These data are summarized in Figures \ref{fig:avg_return} and~\ref{fig:timesteps_reward}.

We observe the trained agent performs 8.8 times better with respect to the reinforcement learning objective than random actions, while it performs on-par with the most proficient baseline. Adding hard-coded heuristics increases return by 2.5 times, making the hybrid policy by far the best performing of those tested. Looking at mean timesteps until reward, we see an opposite trend when introducing the hybrid policy. In this instance, the trained agent performs better than all baselines by over one timestep, while the hybrid policy only averages nearly six more timesteps than the trained agent. These findings highlight that simple hard-coded rules may drastically improve a data-driven policy, but care must be taken to ensure modifications do not have unintended consequences.  

\subsection{Fixation Analysis}\label{subsec:adhoc}

After determining a high-reward vantage point for detecting ripe strawberries, it is optimal for an agent to remain fixated at that location for the remainder of the episode. Fixation behaviors can also be detrimental if they occur on a vantage point with low return. In this section, we seek to characterize the learned agent's fixation tendencies in an attempt to better understand the inner workings of its policy. 

\begin{figure*}[!t]
    \centering 
    \subcaptionbox{high-return fixation}[0.33\textwidth]{\includegraphics[trim={5cm 5cm 5cm 5cm},clip,width=0.33\textwidth]{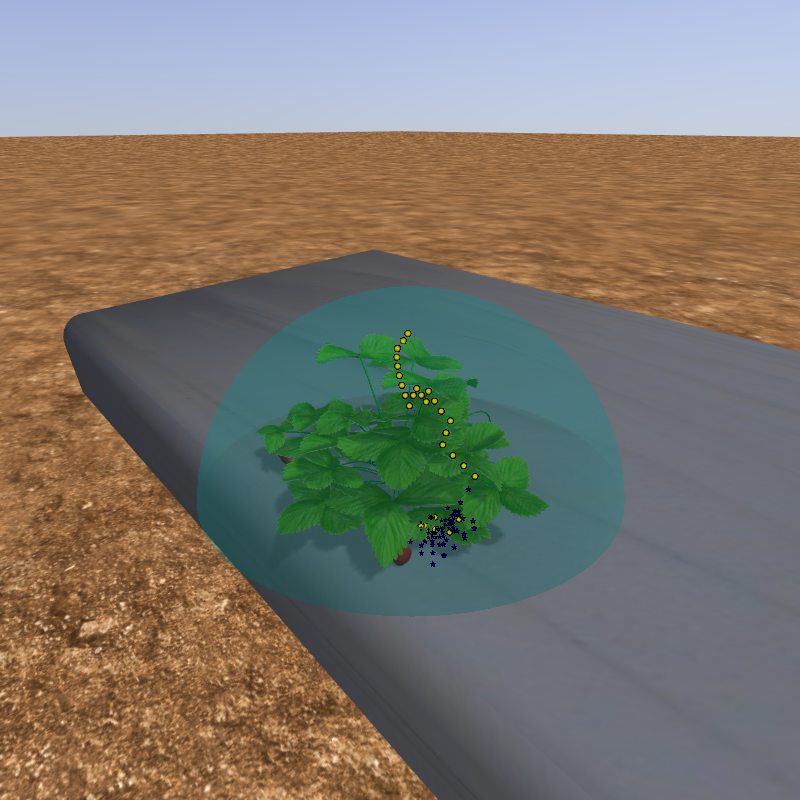}}
    \subcaptionbox{low-return fixation}[0.33\textwidth]{\includegraphics[trim={5cm 5cm 5cm 5cm},clip,width=0.33\textwidth]{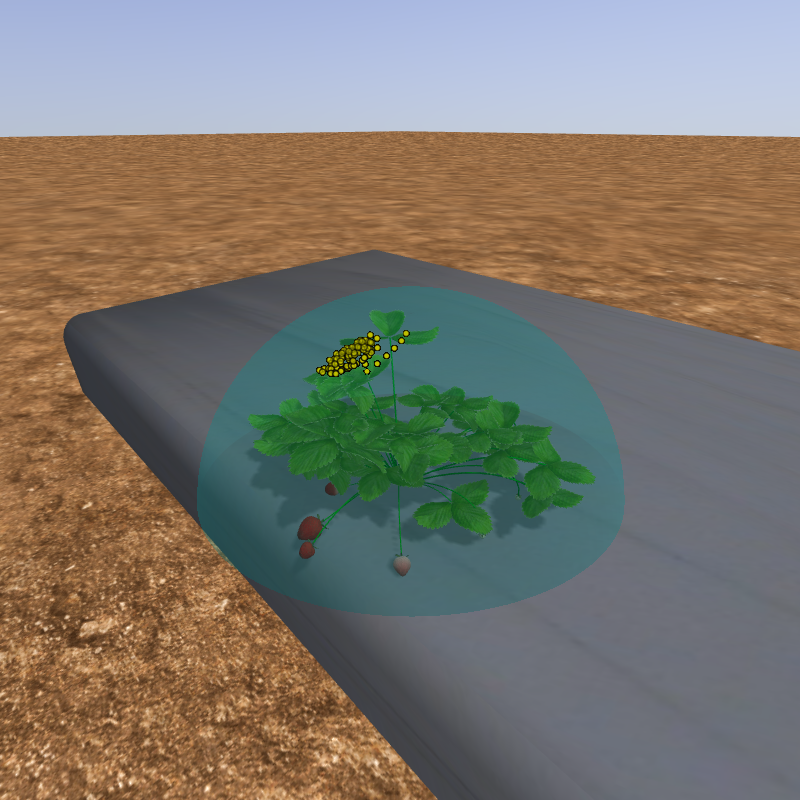}}
    \subcaptionbox{no fixation}[0.33\textwidth]{\includegraphics[trim={5cm 5cm 5cm 5cm},clip,width=0.33\textwidth]{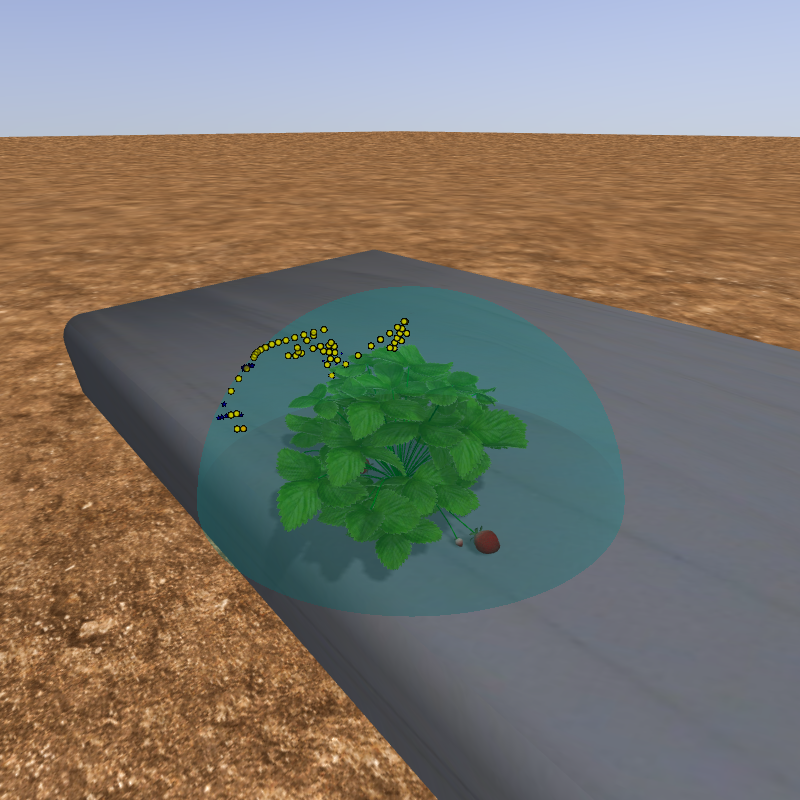}}
    \caption{Sample trajectory visualizations of each category in the simulated environment. Yellow circles represent negative reward, while blue stars indicate a positive detection.}
    \label{fig:visualizations}
\end{figure*}

To detect instances of fixation, we run the DDPG policy on over 400 plants and track positions and rewards for each episode. We plot the corresponding trajectories on a hemisphere superimposed above the plant models, denoting detection states with a blue star. Examples of these visualizations are shown in Figure~\ref{fig:visualizations}.  We then manually inspect the plots and note trajectories where the agent exhibited fixation behaviors.

Of the 213 trajectories with more than 50 steps, 82 fixate on high-return regions, 64 fixate on low-return regions, and 67 do not appear to fixate. From this, it appears the agent learns the desired fixation behavior, but the policy lacks robustness to extend to all states. We suspect some of the low-return clusters are a pitfall of partial observability: Conflicting interpretations of plant geometry at adjacent viewpoints could result in opposite actions for exploration, even if a ripe strawberry is not in frame.  

Using saved plant models from experimental trials, we move the agent to known fixation locations and manually inspect the camera images. On regions with high return, we observe the agent tends to favor viewpoints with ripe strawberries in close proximity with minimal occlusions (Figure~\ref{fig:viewpoint_opt}). Such viewpoints are not only advantageous for detection but likely benefit the remaining steps of the harvesting process, providing better angles for pose prediction and reducing obstacles to simplify planning and execution of harvesting trajectories.

\begin{figure}[!t]
\centering
\includegraphics[width=3.25in]{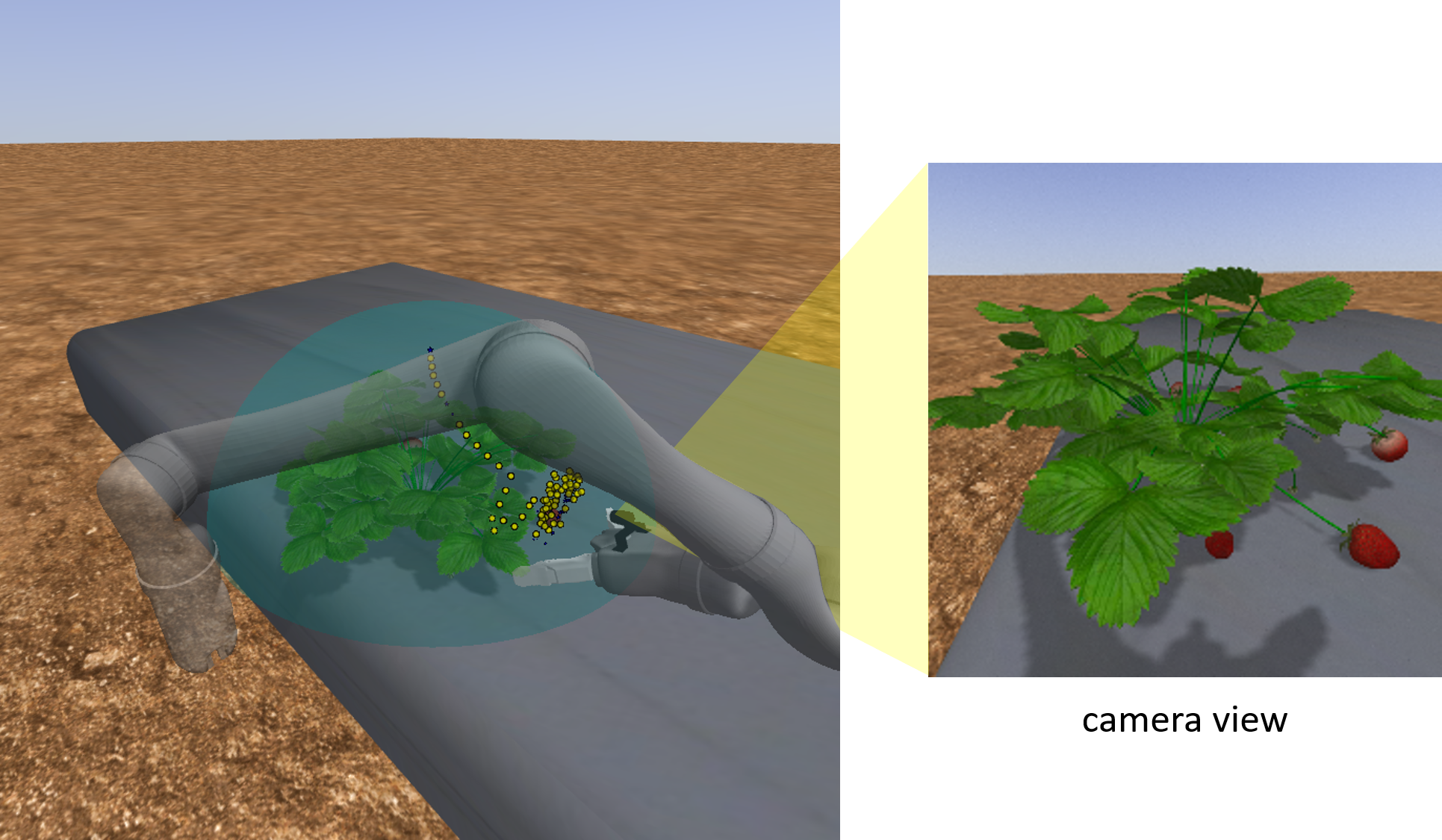}
\caption{\label{fig:first_person_questionable} First-person view from low-return fixation location. Despite yielding low return, ripe strawberries are clearly visible from this vantage point.}
\end{figure}

On the other regions, the image contents are more varied, but we notice that several of the viewpoints display ripe, or nearly-ripe, strawberries in frame that are not picked up by the pretrained detector. An example of this phenomenon is shown in Figure~\ref{fig:first_person_questionable}. In these instances, it is possible that the DDPG policy over-estimates the returns from these viewpoints and thus gravitates towards them. As a whole, it appears that the agent learns to exploit the strengths of the pretrained strawberry detector and generally navigates towards regions where it has a high probability of detection. This is particularly impressive considering the limitations of partial observability. 

\section{Conclusion}\label{sec:conc}

\subsection{Contributions}

In this research, we formulated a novel application of reinforcement learning to solve the viewpoint optimization problem for autonomous strawberry harvesting. We showed that feedback from a pretrained strawberry detector could be used as an autonomous reward scheme, eliminating the need for a human in the loop during the training process. In doing so, we developed a labor efficient data collection procedure for capturing and annotating strawberry images to train the strawberry detector. To train and test our algorithms, we created a realistic simulated environment incorporating many harvesting challenges found in real-world contexts, such as apparent randomness, frequent occlusions, complex textures, and lighting variation.

Within the simulated environment, we saw the agent performed favorably with respect to the reinforcement learning objective and time-to-detection metrics. Additionally, we noted that the nature of the agent's fixated viewpoints not only aids ripe strawberry detection, but also provides desirable angles for pose determination, trajectory planning, and trajectory execution. Therefore, we are optimistic that the viewpoint optimization algorithm would have practical merit in real-world settings. 

\subsection{Future Works.}
The research presented in this paper indicates that reinforcement learning is a promising method to improve the autonomous strawberry harvesting pipeline and serves as a branching-off point for many future developments.  In this section, we briefly outline three recommended directions for future work.

\paragraph{Parallelization.} 
Our training procedure consisted of two independent processes: a worker collecting training data and an asynchronous update procedure using samples from the experience replay. In future work, this could be extended to include multiple workers for the data collection process with minimal modifications to the underlying algorithm. In~\cite{gu2017deep}, parallelization was used to significantly speed up deep reinforcement learning for a real-world manipulation task. Such a modification is a natural extension to autonomous harvesting since most existing systems already use multiple workers during the harvesting process.

\paragraph{Partial Observability.}
In Section~\ref{sec:method}, we defined the agent's ``state" to be its raw camera image and its end-effector position. This presents problems in a heavily occluded environments due to partial observability at each timestep, which may lead to sub-optimal actions. If the agent had some mechanism to propagate relevant state information through time, then it would be able to overcome this limitation. We propose several improvements for future work. First, the state itself could be modified to include additional information from previous state(s). In the simplest case, this could consist of appending the previous position to the current state vector, while a more extreme case is frame stacking. These strategies are limited by their finite horizon and linear memory complexity. A more sophisticated (and likely more successful) approach would be to provide a framework for the agent to learn which information to propagate through time. Possible frameworks for this include recurrent neural networks~\cite{mirowski2016learning}\cite{heess2015memory}\cite{hausknecht2015deep} or an external memory unit~\cite{oh2016control}\cite{zhang2016learning}.

\paragraph{Physical Implementation}
It is imperative to go beyond the simulated environment and evaluate our algorithms on a physical system. While the simulated environment is convenient and enables a rapid development cycle, without physical testing we can only speculate on reinforcement learning's utility in real-world harvesting applications. By comparing results between the two domains, we also will be able to improve the simulated environment and better understand its limitations for future research.

\end{document}